\documentclass[11pt]{article}
\usepackage{coling2020}
\usepackage{times}
\usepackage{url}
\usepackage{latexsym}

\usepackage{algorithmic}
\usepackage{graphicx}
\usepackage{textcomp}
\usepackage{xcolor}
\usepackage{multirow}
\usepackage{amsmath,amssymb,amsfonts}
\usepackage{subfig}
\usepackage{tabularx}
\colingfinalcopy 

\title{Evaluating Bias In Dutch Word Embeddings}

\author{Rodrigo Alejandro Chávez Mulsa\\
  Maastricht University \\
  \texttt{rodrigo.mulsa@outlook.com} \\\And 
   Gerasimos Spanakis \\
  Maastricht University \\
  \\
  \texttt{jerry.spanakis@maastrichtuniversity.nl} \\
  }

\date{31/10/2020}

\begin{document}
\maketitle
\begin{abstract}
 Recent research in Natural Language Processing has revealed that word embeddings can encode social biases present in the training data which can affect minorities in real world applications. This paper explores the gender bias implicit in Dutch embeddings while investigating whether English language based approaches can also be used in Dutch. We implement the Word Embeddings Association Test (WEAT), Clustering and Sentence Embeddings Association Test (SEAT) methods to quantify the gender bias in Dutch word embeddings, then we proceed to reduce the bias with Hard-Debias and Sent-Debias mitigation methods and finally we evaluate the performance of the debiased embeddings in downstream tasks. The results suggest that, among others, gender bias is present in traditional and contextualized Dutch word embeddings. We highlight how techniques used to measure and reduce bias created for English can be used in Dutch embeddings by adequately translating the data and taking into account the unique characteristics of the language. Furthermore, we analyze the effect of the debiasing techniques on downstream tasks which show a negligible impact on traditional embeddings and a 2\% decrease in performance in contextualized embeddings. Finally, we release the translated Dutch datasets to the public along with the traditional embeddings with mitigated bias.
\end{abstract}

\section{Introduction}
\label{intro}

%
%
\blfootnote{
    %
    %
    %
    %
    %
    %
    \hspace{-0.65cm}  
    This work is licensed under a Creative Commons 
    Attribution 4.0 International License.
    License details:
    \url{http://creativecommons.org/licenses/by/4.0/}.
}

In recent years language models have become more relevant in the field of Natural Language Processing (NLP). 
As word embeddings were shown to perform better in many tasks than many traditional techniques, the research community followed this direction and made further advancements resulting in another breakthrough - contextualized word embeddings (e.g. BERT \newcite{Devlin2018BERT:Understanding}, RoBERTa \newcite{Liu2019RoBERTa:Approach}). As opposed to traditional (context-free) word embeddings that have a fixed representation vector, contextualized word embeddings change depending on the sentences (context) in which they are used. They have achieved state-of-the-art in NLP tasks and effectively replaced traditional word embeddings. Due to their outstanding performance, they are broadly adopted in many real-world applications \cite{Wolf2019HuggingFacesProcessing}. 

Models employing those embeddings often support decisions that strongly impact people’s lives, so their fairness and correctness is critical. Unfortunately, previous research has shown that traditional and contextualized word embeddings can encode social biases present in the training data \cite{Caliskan2017SemanticsBiases,Garg2018WordStereotypes,May2019OnEncoders,Zhao2019GenderEmbeddings}.

The social biases in machine learning applications can have an impact on society, as the case in computer vision where three commercial gender classification systems reported higher error rates when recognizing  women, specifically those with darker skin tones \cite{Buolamwini2018Gender}. 
 These biases can cause undesired effects in downstream NLP tasks \cite{Zhao2018IntegratingSimplification,Basta2019EvaluatingEmbeddings} where biased NLP models can amplify bias in real world applications and especially affect minorities which are misrepresented in the data. 
 It has been shown that some minorities like people with disabilities are misrepresented and are associated with a negative sentiment \cite{Hutchinson2020SocialDisabilities}. 

This paper explores the existing gender bias in six traditional and two contextualized Dutch word embedding models by using a combination of state-of-the-art methods proposed in previous literature for English   \cite{Caliskan2017SemanticsBiases,Gonen2019LipstickThem}.
Then, we attempt to mitigate the gender bias in these models and analyze the effect of this step in an applicable downstream task. \cite{Bolukbasi2016ManEmbeddings,Liang2019TowardsRepresentations}
 Therefore we answer the following questions:
 \\
\begin{enumerate}
\item Are Dutch contextualized word embeddings sensitive to gender bias and if so how does the bias compares to traditional Dutch word embeddings?
\item Are measuring and mitigating gender bias methods created for English models compatible with Dutch models?
\item What impact does a bias mitigation method have on the models when used in downstream tasks? 
\end{enumerate}

\section{Related Work}

In social psychology the Implicit Association Test (IAT) \cite{Greenwald1998MeasuringTest} is used to measure the strength of implicit associations between concepts (e.g., black people, gay people) and evaluations (e.g., good, bad) or stereotypes (e.g., athletic, clumsy). When doing an IAT a participant is asked to quickly sort words into categories that are on the left and right hand side of the computer screen. The IAT has five main parts and the score is based on how long it takes a person, on average, to sort the words in the third part of the IAT versus the fifth part of the IAT. We would say that one has an implicit preference for thin people relative to fat people if they are faster to categorize words when Thin People and Good share a response key and Fat People and Bad share a response key, relative to the reverse.

Similarly to the IAT, research in traditional embeddings has proposed how to identify gender bias by measuring the distance between gendered words like pronouns and neutral nouns \cite{Bolukbasi2016ManEmbeddings}, while there has been some relevant work done towards mitigating the bias in the embeddings \cite{Sun2019MitigatingReview}, it has been demonstrated that some of the methods are not enough and the bias can remain hidden within the embeddings. Regardless of the distances in the gender dimension, frequency of words are not taken into account \cite{Wang2020Double-HardMitigation} and biased words can remain clustered together \cite{Gonen2019LipstickThem}. 
Furthermore, these bias methods have to be modified when used on contextualized word embeddings because the embedding representation of every word changes depending on the context in the sentence. 

A modified method is Sent-Debias \cite{Liang2019TowardsRepresentations} which relies heavily on the sentences used to reduce the bias and we hypothesise that due to the extensive use of the pronouns he and she in English, which are not used in Dutch due to their multiple meaning, the mitigation step encompasses a smaller gender subspace in comparison to English and thus the bias is reduced less. 

A predisposition towards English word embeddings exists when researching bias, where the proposed mitigation techniques cannot be directly applied in non-genderless languages. Moreover, there is a lack of work in debiasing contextualized embeddings for other languages that contains characteristics that prohibits to simply import the original methods, like Dutch with \textit{zij} meaning both \textit{she} and \textit{they} in English. While some work has been done in languages that contain grammatical gender like Spanish and French \cite{Zhou2019ExaminingGender}, these proposed methods are restricted to traditional embeddings.

This paper looks into the intersection of the aforementioned techniques by analyzing and mitigating gender bias in traditional word embeddings and the state-of-the-art Dutch models BERTje \cite{DeVries2019}, RobBERT \cite{Delobelle2020RobBERT:Model} while providing a version of the WEAT and SEAT data for this task in Dutch \cite{Caliskan2017SemanticsBiases}.

\section{Bias Statement}
Examples of English NLP models bringing negative consequences to misrepresented groups like an algorithm penalising job applications that contained words like "woman's" and "women’s chess club captain." \cite{Dastin2018AmazonReuters}, or high paying jobs advertisements being shown less to woman than to man \cite{Datta2015AutomatedSettings} are real world cases which could be repeated in Dutch speaking countries, if these Dutch models are deployed without considering the bias present in the word embeddings. In this paper we study the bias by means of embeddings cosine similarities showing that after applying a debiasing technique, other different forms of bias like clustering can remain by having gender stereotype words grouped together enforcing their association. A NLP recommendation system associating words as caring and artistic to girls while using mathematician and sportive to boys could enforce gendered roles and a gender unequal society when for example, recommending toys or books as it has been studied in psychology \cite{Murnen2018FashionBehavior.}.

 The high usability of language models has allowed a broad adoption of these techniques in real world applications, by mitigating different types of bias in NLP models we could not only avoid amplifying these biases but shift the social balance in the long term by avoiding algorithms enforcing social biases against minorities.

\section{Methods and Data}
In the following section we explain the methods and data used in this research. 
Multiple sets of data (e.g. pairs of sentences, lists of gendered words and combinations of sentences of different categories), are needed as a means to analyze and mitigate bias in word embeddings, three algorithms are implemented to measure bias in embeddings; two being applicable to traditional embeddings and the third one is a method adapted for contextualized word embeddings.

Then we show the approaches we use to mitigate the bias in either type of embeddings and test the performance of the bias mitigation on downstream tasks.
Furthermore we recall this research is done on Dutch embeddings, thus the data used is from the same language translated from the English data used on the methods original research.

\subsection{Models}

We perform experiments in six traditional word embeddings, the 300 dimensional Dutch FastText \cite{Grave2018LearningLanguages}, the 320 dimensional small and big embeddings from CLIPS \cite{Tulkens2016EvaluatingResource} trained on Corpora Of the Web (COW), the 160 and 320 dimensional embeddings from CLIPS trained on the Sonar corpus \cite{schafer2012building} and the 100 dimensional Word2Vec from NLPL \cite{Fares2017WordResources}. Furthermore, we test the state-of-the-art 768  dimensional contextualized word embeddings BERTJe \cite{DeVries2019} and RobBert \cite{Delobelle2020RobBERT:Model} which are respectively the homologous versions to BERT \cite{Devlin2018BERT:Understanding} and Roberta \cite{Liu2019RoBERTa:Approach} in Dutch.

\subsection{Bias Measuring Methods}
\subsubsection{Word Embedding Association Test}

The Word Embeddings Association Test (WEAT), as proposed by \newcite{Caliskan2017SemanticsBiases} is a statistical test, similar to the Implicit Association Test (IAT) \cite{Greenwald1998MeasuringTest}, which helps to measure human bias in textual data. 
Both IAT and WEAT use two lists of target words and two lists of attribute words, the first pair of lists correspond to terms we want to compare and the second pair of lists represent the categories in which we believe bias can be present. 

\newcite{Caliskan2017SemanticsBiases} defined ten tests using WEAT to measure the bias in different categories\footnote{Full list in Appendix including extra tests 11-16.}

In our research we translate the WEAT lists of words used in the tests to Dutch and modify them accordingly so words in these lists remain associated only to the corresponding category. Some of the modifications correspond to the different linguistic characteristics of the language and the lack of meaningful translations of certain words in the data (e.g. avoid using \textit{zij/ ze} which they both can be used as what would be \textit{she} and \textit{they} in English making them not candidate words to represent gender since they can encode neutral plural too). We put special attention to Weat- \textit{6, 7} and \textit{ 8} since these tests measure the gender bias in three different lists\footnote{The corresponding terms in Dutch are: \textit{carriere vs gezinsactiviteiten, wiskunde vs kunst, and wetenschap vs kunst.}} of attributes; Career vs Family, Mathematics vs Art, and Science vs Art. \textit{Weat-6} uses male and female names used in English\footnote{We created an alternative list of target words for seat-6 with a list of dutch female and male names, based on popularity retrieved from the Sociale Verzekeringsbank
\textit{(Social Insurance Bank in English)}, results are shown in the appendix.} while \textit{weat 7 and 8} use gendered words for the target lists (e.g. grandpa, grandma). 
WEAT consists of two parts, the p-values ($p$) and the effect size ($d$). The test statistic \eqref{inequality_weat}  indicates the significance where a p-value larger than $0.05$ indicates the bias is insignificant, while the effect size \eqref{effect-size} measures the magnitude of the associations representing how much bias is quantified. 

With the purpose of measuring bias, 
we use the WEAT test statistic that measures the difference of the aggregated similarities between a target word list, with notation $M$ or $F$, to the attribute word lists $A$ and $B$. The target to attribute similarities are computed by getting the mean of the cosine similarity between the words in a target list (e.g. $M$) and the words in an attribute list (e.g. $A$) and then subtracting the mean of the same measurement of the second attribute list (e.g. $B$).  The test statistic can be described as: 

\begin{equation}
s(M,F,A,B) = [\sum_{m\epsilon M}^{} s(m,A,B) - 
\sum_{f\epsilon F}^{} s(f,A,B)]  \label{WEAT}
\end{equation} 

where $s(w, A, B)$ corresponds to the cosine similarities between some word $w$ and the attribute words $a$ and $b$ represented as: 
\begin{equation}s(w,A,B) = [\frac{\sum_{a\epsilon A}^{}cos(w,a)}{|A|}-\frac{\sum_{b\epsilon B}^{}cos(w,b)}{|B|}] \label{eq}
\end{equation}

In order to compute the significance we first merge the $M$ and $F$ lists, then we generate $100,000$\footnote{If there are more than $100k$ possible permutations, we sample $99,999$ and assume at least one of the permutations satisfied the inequality in \eqref{inequality_weat} to account for the loss of precision.}  permutations, denoted as $PERM$, of this combined list and split each of them in new pairs of $M_{i}$ and $F_{i}$ lists, we perform the test statistic \eqref{WEAT} on every pair and calculate the test significance by looking at the amount of permutations where the result of the test statistic is higher than the result of the original tested lists, and divide this count by the total amount of permutations (single tail test). Described as:

\begin{equation}
p = \dfrac{\sum^{}_{i\epsilon PERM} [s(Mi,Fi,A,B)>s(M,F,A,B)]}{|PERM|} 
\label{inequality_weat}
\end{equation}

Furthermore, the amount of bias in WEAT is analyzed using the effect size \textit{d} that is computed by obtaining the difference between the mean of the cosine similarities between the $M$ and $F$ lists to the attributes $A$ and $B$, and normalizing them by dividing them with the standard deviation of both lists combined, which can be formulated as:

\begin{equation}
    \begin{aligned}
          d = \dfrac{mean_{m\in M} s(m,A,B)-mean_{f\in F} s(f,A,B)}{\mathrm{stddev}_{w\in M\cup F}s(w,A,B)}
    \end{aligned}
\label{effect-size}
\end{equation}

\subsubsection{Clustering accuracy}
A different metric was introduced by \newcite{Gonen2019LipstickThem} showing that word embeddings with mitigated bias can remain clustered together even though the distance between attribute and target words (in WEAT) is insignificant. 
The clustering accuracy test requires to project the whole vocabulary into a male and female term to get the gender direction of each word in the vocabulary. \newcite{Gonen2019LipstickThem} used the pronouns \textit{he} and \textit{she} because they are widely used and the only difference between them is in the gender subspace.
The homologous pronouns in Dutch are \textit{hij} and \textit{zij} which represent a problem in this research due to \textit{zij} meaning both \textit{she} and \textit{they} thus adding extra meaning besides gender to the geometrical difference of the pronouns.
We tested different\footnote{Some of the candidate pairs include [\textit{jongen, meisje}] \& [\textit{mannelijk, vrouwelijk}]. } pairs of words that could better substitute the pronouns in the projection step and selected the pair \textit{man} and \textit{vrouw} to generate the gender direction of the words in the vocabulary.
We implement this test for traditional word embeddings, first we compute the gender direction of every\footnote{We limit the vocabulary in this test by skipping words where gender gives them part of its meaning (e.g. \textit{grandma}/\textit{grandpa}). } word in the vocabulary $W$ by getting the dot product of each word to a male and female word, 'man' and 'vrouw' in this case, and getting the difference of the scalars as the gender direction of each word. Then we sort the entire vocabulary based on this gender direction and retrieve the $k$ most associated to the male and female direction\footnote{In our experiments we get the $500$ most female and the $500$ most male oriented words.}. 

We use this top $2k$ most biased words and create ground truth labels $l\in{L}$ where $l=0$ if the word is from the $k$ male words and $l=1$ if the word is from the $k$ female words. Next we run KMeans predictions to categorize the words giving to each word a predictive label $\hat{l}$.

\begin{equation}
 a = \frac{1}{2k} \sum_{i=1}^{2k} 1[l_{i}==\hat{l_{i}} ] \label{clustering}
\end{equation}

We use \eqref{clustering} to count how many of the words were clustered correctly according to the ground truth labels $l$ and compare this metric with the traditional word embeddings before and after mitigating the bias.  Then, set $a=max(a,1-a)$ to test if the words remain clustered after mitigating the bias. Given $a$, the closer it is to $1$ the more biased the embeddings are, while a more random clustering is represented the closer its value is to $0.5$, thus the least bias there is.

\subsubsection{SEAT}

Following from \newcite{Caliskan2017SemanticsBiases}, \newcite{May2019OnEncoders} proposed a method adapted from WEAT that can be used in contextualized word embeddings like BERT, by converting every word in WEAT into multiple sentences using a set of semantically bleached sentence templates. Then the same formulas are used as in WEAT where the embeddings represent the entire sentence instead of only a word. This approach hypothesizes that the models that use context to get more accurate vector representations should not be tested on a word basis like WEAT, and by converting the original WEAT lists of words into sentences with multiple contexts the models can generalize better and therefore can be tested too for biases.

Similar to the WEAT lists of words, we translated and adjusted the sentences used in SEAT to Dutch. \newcite{May2019OnEncoders} enriched the tests by adding \textit{b} versions of the tests 3, 5, 6, 7 and 8 by replacing given names with group terms (e.g., male, son, female, daughter) and vice versa. Similar as with WEAT, in our results we focus on the tests 6, 6b, 7, 7b, 8 and 8b which are related to gender bias. 

Examples of the sentences templates introduced by \newcite{May2019OnEncoders} include: 'This is a \textless \textit{WeatWord}\textgreater', 'That is a \textless \textit{WeatWord}\textgreater'. We adapted all the SEAT templates to Dutch and this equivalent sentences become:  'Dit is een \textless \textit{WeatWord}\textgreater', 'Dat is een \textless \textit{WeatWord}\textgreater'. 

\subsection{Debiasing Methods}
\subsubsection{Hard-Debias}
It was demonstrated with WEAT that there is bias in traditional word embeddings when comparing attributes and target words. Based on the WEAT mathematical definition of bias, \newcite{Bolukbasi2016ManEmbeddings} hypothesized there is a gender direction encoded in these embeddings which can be subtracted from the word representations and by equalizing the distance between similar terms (e.g., \textit{he} and \textit{she}) it is possible to reduce the bias in the word embeddings.

Hard-Debias requires three lists of words, a gender specific list, a definitional list and an equalize list. The first list contains a broad list of gender related words that are not debiased since their meaning depends at least partially on the gender subspace, the second list correspond to the words used in the PCA to define the bias subspace and the third list contains words that should be equalized in opposite direction in the gender subspace with the same magnitude. We translated to Dutch the lists used in Hard-Debias and modified them similarly as with the previous data to adjust for the language differences. 

Following the description by \newcite{Wang2020Double-HardMitigation}, given a vocabulary $W$ from some word embeddings, we can denote each word embedding in $W$ as $\overrightarrow{w} \in \mathbb{R}^{n}$ per each  $ w \in{W}$. We define a subspace in $W$ as $B$ which is defined by $k$ orthogonal unit vectors $B = \{b_{1},...,b_{k}\} \in \mathbb{R}^{n}$. We can describe a projection of some embedding $\overrightarrow{w}$ on $B$ by:

We assume a predefined set of $n$ (male, female) pairs of words  $D_{1},D_{2},...,D_{n} \subset  W$, where the main difference between each pair of words is the gender \cite{Bolukbasi2016ManEmbeddings}. Let

\begin{tabularx}{\textwidth}{Xp{2cm}X}

\begin{equation}
\overrightarrow{w}_{B}=\sum_{j=1}^{k} (\overrightarrow{w} \cdot  b_{j})b_{j}
\end{equation}

& &
\begin{equation}
    \mu _{i} := \sum_{w \in D_{i}}\vec{w}/|D_{i}|
    \label{meanHD}
\end{equation}

\end{tabularx}



Then we identify a gender subspace $B$ that captures the gender bias. Following from \eqref{meanHD} we can compute $B$ as the first $k \geq 1$ components from the principal component analysis (PCA) \cite{Abdi2010PrincipalAnalysis}.

\begin{equation}
B = PCA_{k}(\bigcup^{n}_{i=1} \bigcup_{w\in{D_{i}}}(\vec{w}-\mu_{i} ))
\end{equation}
Then we neutralize the word embeddings corresponding to words that are neutral to gender by modifying each $\vec{w} \in \mathbb{R}^{n}$ such that every word $w \in N$ has a zero projection in the gender subspace. 
\begin{equation}
\vec{w}=\vec{w}-\vec{w_{B}}
\end{equation}
Finally we equalize the gendered word pairs $D_{i}$ (e.g. \textit{man} and \textit{vrouw}) such that they have a gender component in opposite directions but with the same magnitude. This ensure that the distance between neutral words to biased words is the same with respect to the bias subspace.

\subsubsection{Sent-Debias}
The added context in sentence embeddings and the unlikelihood of retraining models due to their increasing size like GPT2 \cite{Radford2019LanguageLearners}, have made debiasing of these models even harder.

A modified Hard-Debias compatible with sentence representations was proposed by skipping the equalizing step from the original method, and studying the effect and performance of diverse sentence pairs that encode the bias subspace \cite{Liang2019TowardsRepresentations}. 
We generate a new dataset of 30,000 gendered sentence pairs in order to compute the gender direction by using the Wikipedia Monolingual Corpora in Dutch from Linguatools \cite{Kolb2018WikipediaLinguatools} and the Gensim library \cite{Sojka2010SoftwareCorpora}. After processing the corpora into single sentences\footnote{With a maximum length equal to the longest length of accepted inputs in BERTJe and RobBert (512).} we proceed to find $k$ amount of sentences that contains one of the words from the gendered word list used in Hard-Debias, we save the sentence as a tuple with the new sentence created by swapping the gendered word in the original sentence by the same word of opposite gender from the gendered list, making them different only in the bias subspace (e.g. "\textit{She} is the best friend of Obama" creates the pair: "\textit{He} is the best friend of Obama"). 

We compute the gender subspace and mitigate the bias by making the sentence embedding orthogonal to this gender subspace.  A major consideration we had to take when creating this Dutch dataset was that in this language, some of the pronouns that are gender related, like \textit{zij/ze}, can be associated to the female gender or be used as plural (\textit{she/they}), while the male pronoun \textit{zij} is also used as a subjunctive with every pronoun, thus an unsupervised method to generate the dataset cannot include these words since it would bring noise to the subspace we want to generate. Similar as in Hard-Debias we use these pairs of sentences with the principal component analysis (PCA) to get the gender subspace made of the first $k$ components of the PCA. We then project each sentence embedding into each of the components and use the sum of these vectors to subtract the gender subspace from every sentence embedding to mitigate the bias.


    


\subsection{Downstream tasks}
\subsubsection{Relation Identification Task}

Data for this Dutch task was created following a similar task used in the original \textit{word2vec} toolkit\footnote{https://code.google.com/archive/p/word2vec/} which contains approximately 20,000 relation identification questions, each of the form: “If A has a relation to B, which word has the same relation to D?. This Dutch dataset created by \newcite{Tulkens2016EvaluatingResource} aims to replicate the original evaluation set as closely as possible, while also including some characteristics of Dutch that are not present in English, such as the formation of diminutives, and thus being a more accurate evaluation task for the models in Dutch.

The task is described as approximately 20,000 relation identification questions, each of the form: “If A has a relation to B, which word has the same relation to D?”. As such, it uses the fact that vectors are compositional. For example, given man, woman, and king, the answer to the question should be queen, the relation here being ‘gender’. 

\subsubsection{Sentiment Analysis Task}

In most real world applications the contextualized word embeddings will be fine-tuned to a specific task in which they are used, we proceed to finetune the models BERTJe and RobBERT, on the Sentiment Analysis task \cite{vanderBurgh2019TheReviews} using the 110k Dutch Book Reviews dataset (DBRD)\footnote{Dataset available at: \\ \url{https://github.com/benjaminvdb/110kDBRD}} and test the performance difference when the model bias has been mitigated. The dataset has been used to compare the original performance of BERTJe and RobBERT, it is split in a balanced 10\% test (2224 reviews) and 90\% train split where each review is labeled as positive or negative.  

\begin{table*}[htb!]

\begin{center}
\small
\begin{tabular}{|c|c|c|c|}
\hline
WEAT test & Weat-6                                    & Weat-7                                    & Weat-8                                  \\ \hline
FastText  & $1.534^{**}  \rightarrow 1.605^{**}$          & $1.484^{**}  \rightarrow \textbf{1.260}^{**}$ & $1.147^{**}  \rightarrow \textbf{0.672}$ \\ \hline
COW-small & $1.866^{**}  \rightarrow \textbf{1.840}^{**}$  & $1.759^{**}  \rightarrow \textbf{0.947}^{*}$  & $1.339^{**}  \rightarrow \textbf{0.392}$  \\ \hline
COW-big   & $1.771^{**}  \rightarrow \textbf{1.738}^{**}$ & $1.713^{**}  \rightarrow \textbf{1.100}^{*}$  & $1.425^{**}  \rightarrow \textbf{0.506}$ \\ \hline
Sonar-160 & $0.726  \rightarrow \textbf{0.578}$     & $1.451^{**}  \rightarrow \textbf{0.410}$   & $1.180^{**}  \rightarrow \textbf{0.271}$ \\ \hline
Sonar-320 & $0.528  \rightarrow \textbf{0.526}$     & $1.1716^{*}  \rightarrow \textbf{0.643}^{*}$   & $0.995^{*}  \rightarrow \textbf{0.615}$  \\ \hline
Model-NLPL & $1.748^{**}  \rightarrow \textbf{1.721}^{**}$ & $1.443^{**}  \rightarrow \textbf{1.161}^{**}$ & \multicolumn{1}{c|}{$0.766  \rightarrow 0.885^{*}$}  \\ 
\hline 

\multicolumn{4}{l}{Arrow indicates before to after mitigating bias; * indicates significant at 0.05, ** significant at 0.01.}
\end{tabular} 

\caption{WEAT effect size on gender related test.}
\label{weat-results}

 \end{center}
\end{table*}

\section{Experiments and Results}

In this section we explain how we use the methods and data explained in previous sections to measure and mitigate the bias in Dutch embeddings. We demonstrate the existence of bias in traditional word embeddings and explain the bias mitigation process that has been used. Then we demonstrate that contextualized word embeddings show less bias than traditional ones while also determining that the bias mitigation is not as effective in contextualized word embeddings as in traditional word embeddings.
\subsection{Results On Word Embeddings}
\subsubsection{WEAT}

We performed the WEAT test  on the 16\footnote{All the results are available in the appendix.} adapted lists of words translated to Dutch. Among all the traditional word embeddings we see high effect sizes and multiple tests are significant at different levels. 

The results of the WEAT effect sizes on gendered related tests are shown in table \eqref{weat-results} where we see an overall high effect size across all the scores on the original models. Then we look into the results after the debiasing step (at the right of the arrow) where it shows that the bias mitigation is effective in almost every test for every model with just 2 cases where the score did not decrease. If the gender related results of the FastText model are compared to the rest of the tests in the appendix \eqref{FasttextWEAT} for example, we can see how other types of bias are present, but the mitigation step mainly provides good results in the gender related tests by reducing the gender bias further than on other types of biases.

In the other models we also see a high amount of significant tests where the bias mitigation has a positive impact on every test effect size and some of the bias loses its significance. We note the models by Clips (COW and Sonar) contain significant bias with a relative high effect size specially in tests with gender words as target  ($d>1.0$).  Furthermore, the results in table \eqref{weat-results} for the Word2Vec NLPL model are also high and significant on Weat-6 and Weat-7. 

\subsubsection{Clustering Accuracy Test}

We perform the clustering accuracy test to compare if biased words remain clustered together even after performing the Hard-Debias method to mitigate the gender bias present in the embeddings. Our results matches the English results from  \newcite{Gonen2019LipstickThem} which show that mitigating bias focusing only on WEAT can hide bias which is measured differently. On table \ref{table-clustering}, using formula \eqref{clustering} we compare the accuracy in which KMeans creates two clusters using the previously defined biased words before and after mitigating the bias. Since the goal is to have embeddings where there are no biased words, attempting to cluster the words should perform randomly and not create groups of male and female words, thus the closer the score to $0.5$ the better. 

\begin{table}[htb!]

    \begin{center}

\small
    \begin{tabular}{|c|c|c|}
    \hline 
        Model &  Original &  Debiased \\
       \hline
    FastText & 0.611 & 0.605 \\
 \hline
 COW-small & 1.0 & 1.0 \\
 \hline
 COW-big & 0.999 & 0.999 \\
 \hline
 Sonar-160 & 1.0 & 1.0 \\
 \hline
 Sonar-320 & 0.998 & 0.998 \\
 \hline
 Model NLPL & 0.999 & 0.995 \\
 \hline
 
\multicolumn{3}{l}{Before and after the bias mitigation step.}
  \end{tabular}
\caption{Cluster test results}

\label{table-clustering}     

\end{center}
\end{table}

The results in \eqref{table-clustering} show that after mitigating the bias on the FastText model the accuracy slightly decreases by $0.006$ hence the Hard-Debias method can work in reducing the cluster bias, while the results also show there is no change in the predictions on the Clips models having an almost perfect accuracy, and in the case of the NLPL model the score decreases slightly by $0.004$ like in the FastText case.  Our results in this Dutch models are similar to the ones presented in English models \cite{Gonen2019LipstickThem} which demonstrates that bias can be barely unaffected by some debiasing methods when analyzed with a different metric than WEAT.

\begin{figure}
    \centering
\includegraphics[width=0.60\textwidth]{"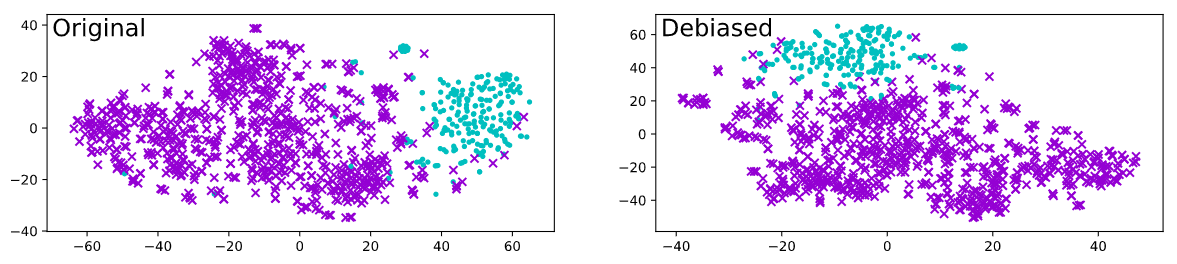"}
\caption{Example of clustered bias on FastText in dutch. }
    \label{fig:fasttext}
\end{figure}
 
 We also provide an example in figure \eqref{fig:fasttext} of the 500 female and 500 male most biased words clustering together before and after the mitigation step, showing that even though their position changes, they remain grouped together. 

%

\subsubsection{SEAT}

We use the Huggingface transformers library  \cite{Wolf2019HuggingFacesProcessing} to load and use both BERTJe \cite{DeVries2019} and RobBert \cite{Delobelle2020RobBERT:Model}. To retrieve the sentence embeddings we use the special tokens corresponding to each model, in BERTJe we use the [CLS] token while in RobBert we use the $<s>$ token. 

We then load the bare models to compute the adapted SEAT test to BERTJe and RobBERT. In BERTJe we see that the tests like SEAT-1\footnote{Available in the appendix.} has no significant bias but the gender oriented tests like SEAT-6, 6B, 7b and 8 show bias; although the effect size is less compared to the previously shown results in the word embeddings. This has been hypothesized as contextual embeddings are less susceptible to bias due to the context words influence, yet the results have proved these embeddings are not exempt of bias and a bias mitigation technique is needed.


We then modify the Huggingface transformers library to implement the mitigation step every time we generate an embedding. The following table \eqref{SEAT_results} shows the SEAT results, where the bias mitigation applied using Sent-Debias in both models, reduce the effect size in four out of six tests, but the change on the bias scores in this metric is minimal compared to the bias mitigation in traditional embeddings. This could be due to contextualized embeddings encoding the gender direction along with the bias in a more complex subspace than the one used in traditional embeddings.

\subsection{Performance On Downstream Tasks}
\subsubsection{Relation Identification Task}

We test the embeddings before and after mitigating the bias on a relation identification task in Dutch. 
We present in our results the accuracy in which the models find a word $w$ in the vocabulary $W$  which maximizes the similarity score with the vector $(A - B + D)$.

Our results are displayed in table \ref{downstream-weat} having mixed scores after the bias mitigation step where the biggest amount decreased is shown with the smallest model Sonar-160 by $0.01397$ and the models Cow-big and NLPL show a small increase in their scores by $0.00225$ and $0.00305$ respectively.

\subsubsection{Sentiment Analysis Task}

For this task we report the accuracy score, the 95\% confidence intervals\footnote{Confidence intervals are calculated using the normal approximation to the binomial distribution.} and F1 score. The results displayed in table \eqref{downstream-seat} show that both of the models performance decreases approximately 2\% in the downstream task after going through the bias mitigation, which correlates to similar results that have been presented in previous research \cite{Liang2019TowardsRepresentations}. We hypothesize this could be due to the transformation the embeddings go through when mitigating the bias or because the dataset used in the downstream task is biased thus conflicting with the model training. \begin{table}[!htb]
\begin{minipage}[t]{0.48\linewidth}
    \centering
\medskip
\begin{center}
\small

    \begin{tabular}{|c|c|c|}
    \hline
        SEAT list &  BERTJe &  RobBERT \\ \hline
 SEAT-06 & $0.666^{**}  \rightarrow \textbf{0.614}^{**}$ & $0.191  \rightarrow \textbf{0.190}$ \\
 \hline
 SEAT-06b & $0.349^{*}  \rightarrow \textbf{0.294}^{*}$ & $0.552^{**}   \rightarrow 0.5846^{**} $ \\
 \hline
 SEAT-07 & $0.6300^{**}  \rightarrow \textbf{0.588}^{**}$ & $0.331^{*}  \rightarrow \textbf{0.252}$  \\
 \hline
 SEAT-07b & $0.613^{**}  \rightarrow \textbf{0.605}^{**}$ &$0.683^{**}  \rightarrow \textbf{0.669}^{**}$  \\
 \hline
 SEAT-08 & $0.089  \rightarrow 0.114$ &$0.080  \rightarrow 0.105$ \\
 \hline
 SEAT-08b & $0.286  \rightarrow 0.294$ & $0.157  \rightarrow \textbf{0.149}$ \\
 \hline

\multicolumn{3}{l}{Arrow indicates before to after mitigating bias; * indicates}\\
\multicolumn{3}{l}{ significant at 0.05, ** significant at 0.01.}
 \end{tabular}
     \caption{SEAT effect size on gender related tests.}

\label{SEAT_results}
\end{center}
\end{minipage}\hfill
\begin{minipage}[t]{0.48\linewidth}
    \centering
\medskip

    \begin{center}
\small
    \begin{tabular}{|c|c|c|}
    \hline 
        Model &  Original &  Debiased \\
       \hline
    FastText-320 & 0.67445 & 0.662 \\
 \hline
 Cow-320 & 0.513 & 0.512 \\
 \hline
 Cow-Big & 0.516 & 0.518 \\
 \hline
 Sonar-160 & 0.408 & 0.394 \\
 \hline
 Sonar-320 & 0.429 & 0.421 \\
 \hline
 Model-NLPL & 0.430 & 0.433 \\
 \hline
\multicolumn{3}{l}{Before and after the bias mitigation step.}
  \end{tabular}
\caption{Relation identification task results}

\label{downstream-weat}
\end{center}

%
\end{minipage}
\end{table}

\begin{table*}[htb!]

    \begin{center}
\small
    \begin{tabular}{|c|c|c|c|c|}
    \hline 
    \multirow{2}{*}{Model} &  \multicolumn{2}{|c|}{Original}  & \multicolumn{2}{|c|}{Debiased} \\
    \cline{2-5}
     & ACC (95\% CI) [\%] & F1 [\%] & ACC (95\% CI) [\%] & F1 [\%] \\
       \hline
    BERTJe & 0.952 (0.943,0.961)&0.952	
	&  0.946	(0.937,0.955)	 &0.946	

\\
 \hline
 RobBERT & 0.935 (0.925,0.945) &0.935
 & 0.913  (0.901,0.925) &0.913

\\
 \hline
 
\multicolumn{5}{l}{Before and after the bias mitigation step.}
  \end{tabular}

     \caption{Sentiment analysis task results}
\label{downstream-seat}
\end{center}
\end{table*}

\section{Discussion}
In the traditional word embeddings we see large effect sizes ($d>1.0$) in every WEAT test that passes the significance statistic and we manage to decrease these scores with the Hard-Debias method. We also note that the bias mitigation on tests that uses names in either English or Dutch doesn't perform as well as in the other tests that terms of groups are used. We hypothesize that while names can encode bias, it could be in a different subspace than the group terms and if not enough names are used in the data of Hard-Debias to generate the gender subspace then the mitigation fails when dealing with names.

We point out that Hard-Debias has an equalizing step that depends in a list of gendered word pairs which should not be transformed by the bias direction and instead centralized where each pair is separated by the same distance. This can be a limitation on the method since it heavily relies in this list of gendered word pairs where having an incomplete list could translate into bad performance of the model if words that depend on the gender dimension for their meaning lose it.

The clustering metric shows that even if the bias is reduced using Hard-Debias and tested on WEAT, there can be other approaches to identify bias and some methods won't mitigate the bias from that point of view, therefore we suggest more research should be done in different ways to identify bias and mitigate it.

The results on the relation identification task showed that the bias mitigation step had a slight influence on the model results, but overall the method shows that Hard-Debias does not affect negatively in a remarkable amount the performance on the task. 

As for the results of BERTJe and RobBert, they both show bias in the same tests associated to gender but present a smaller effect size overall ($d < 1.0$) compared to traditional embeddings on WEAT. When using Sent-Debias the effect sizes decreases along the significance statistic of the bias, yet the tests remain in the same statistical significance category ($p\leq0.05$ $\vee$  $p\leq0.01$). This shows that the bias mitigation technique works but it does not completely eliminate the bias. A possible reason to this result is that contextualized words embeddings contain a more complex gender subspace and a stronger method might be necessary in order to mitigate the bias on it.

Moreover, we show on the sentiment analysis task that similarly to the English research \cite{May2019OnEncoders}, when mitigating the bias the accuracy on downstream tasks decreases by a small percentage, this demonstrates that methods built for English embeddings can also be used for Dutch ones and produce similar results on equivalent tasks. 

One of our work's limitations is that while our methods are directly applied in Dutch, our test data was created in English and adapted for Dutch. That means that our tests do not specifically take into consideration some of the linguistic features in Dutch that are not present in English, such as the use of diminitives (e.g. \textit{bier} $\rightarrow$ \textit{biertje}) or the distinct styles of expressions in the Dutch dialects, regional languages (e.g. \textit{Limburgish}) or the Afrikaans sister language.

We also highlight that both WEAT and SEAT don't test for the lack of bias, just test whether if in the test cases bias exist but there could be other non-tested cases where bias is present. An example of this is shown when computing the clustering accuracy test which demonstrates that if we measure bias from a different point of view, the bias remains and thus further research in different bias mitigation techniques is recommended.

\section{Conclusion}
With this research we show that among others, gender bias is present in Dutch traditional and contextualized word embeddings (as probably in other languages as well). We then show how methods used to measure and mitigate bias in English embeddings, can be used in Dutch embeddings by properly translating the data and taking into consideration the unique characteristics of the language (\textit{e.g. zij/ze}). Furthermore, we analyze the effect of the mitigating techniques in downstream tasks showing negligible impact in traditional embeddings and approximately a 2\% decrease of performance in contextualized embeddings, which can be considered detrimental if the embeddings used guarantee a more gender-neutral approach.

A promising future direction stemming from this research would be the development of extensive bias tests taking into consideration Dutch dialects and other language peculiarities. Moreover, more research is needed into the evaluation of different definitions of bias in contextual embeddings as we have shown in the findings that clustering bias persists regardless of the WEAT results in traditional embeddings. With this paper, we highlight the existence of bias in commonly used representation models, therefore, we advocate the use of mitigated bias models, especially in many industry applications (e.g. real-world NLP models) where bias existence can harm minorities.

Finally, we release the new Dutch datasets for all bias tests to the public\footnote{Available at: \url{https://github.com/Noixas/Official-Evaluating-Bias-In-Dutch}} and the traditional embeddings with mitigated bias.

\section*{Acknowledgement}
Authors would like to thank Visma Connect B.V. for providing the cloud resources that allowed carrying out this research and more specifically Yvo Keuter for his supervision and mentorship during Rodrigo's time at Visma Connect.

\bibliographystyle{coling}
\bibliography{references,coling2020}

\appendix

\section{Appendices}

\label{sec:appendix}

\subsection{WEAT and SEAT Lists}

In table \eqref{weat_categories} we show each of the tests and categories used  in WEAT. Where for example in Weat-1 we test for bias in pleasant and unpleasant words associated to insects and flowers.

\begin{table}[hb!] 
\caption{WEAT examples lists of words}
    \begin{center}
   \small
\begin{tabular}{llll}
\hline
  & Category      & English     & Dutch          \\ \hline
0 & family        & home        & thuis          \\
1 & family        & parents     & ouders         \\
2 & family        & children    & kinderen       \\
3 & family        & family      & familie        \\
4 & family        & cousins     & neven          \\
5 & family        & marriage    & huwelijk       \\
6 & family        & wedding     & bruiloft       \\
7 & family        & relatives   & familieleden   \\
0 & math          & math        & wiskunde       \\
1 & math          & algebra     & algebra        \\
2 & math          & geometry    & geometrie      \\
3 & math          & calculus    & calculus       \\
4 & math          & equations   & vergelijkingen \\
5 & math          & computation & berekening     \\
6 & math          & numbers     & getallen       \\
7 & math          & addition    & optellen       \\
0 & arts          & poetry      & poëzie         \\
1 & arts          & art         & kunst          \\
2 & arts          & dance       & dans           \\
3 & arts          & literature  & literatuur     \\
4 & arts          & novel       & roman          \\
5 & arts          & symphony    & symfonie       \\
6 & arts          & drama       & drama          \\
7 & arts          & sculpture   & beeldhouwwerk  \\
0 & male\_terms   & male        & mannelijk      \\
1 & male\_terms   & man         & man            \\
2 & male\_terms   & boy         & jongen         \\
3 & male\_terms   & brother     & broer          \\
4 & male\_terms   & he          & hij            \\
5 & male\_terms   & him         & hem            \\
6 & male\_terms   & his         & zijn           \\
7 & male\_terms   & son         & zoon           \\
0 & female\_terms & female      & vrouwelijk     \\
1 & female\_terms & woman       & vrouw          \\
2 & female\_terms & girl        & meisje         \\
3 & female\_terms & sister      & zus            \\
4 & female\_terms & she         & zij            \\
5 & female\_terms & her         & haar           \\
6 & female\_terms & hers        & haar           \\
7 & female\_terms & daughter    & dochter        \\ \hline
\end{tabular}
\end{center}
\end{table}

\begin{table*}[htb!]
    \begin{center}
    \small
    \begin{tabular}{|c|c|c|}
    
    \hline
     Category &  English &  Dutch \\
     \hline
Math & This is an equation. & Dit is een vergelijking. \\
       \hline
Math & That is an equation. & Dat is een vergelijking. \\
       \hline
Arts & This is a sculpture. & Dit is een sculptuur. \\
       \hline
Arts & That is a sculpture. & Dat is een sculptuur. \\
       \hline
MaleTerms & The brother is here. & De broer is hier. \\
       \hline
MaleTerms & The brother is there. & De broer is daar. \\
       \hline
FemaleTerms & A woman is a person. & Een vrouw is een persoon. \\
       \hline
FemaleTerms & The woman is there. & De vrouw is daar. \\
       \hline
  \end{tabular}

    \caption{Examples of sentences used in SEAT-7, English and Dutch.}
\label{seat-7-examples}
\end{center}
\end{table*}

\subsection{Full WEAT tests results}
In the following tables we show the results of the 16 WEAT tests per model, indicating the precise effect size and p-value per test.

\begin{table}[htb!]
     
\begin{minipage}[t]{0.48\linewidth}
\centering
\medskip
     
    \begin{center}
\small
    \begin{tabular}{|c|c|c|}
    \hline
        WEAT list &  Effect size d &  Significance p \\
       \hline
    Weat-1 & $1.376  \rightarrow 1.416$ & $0.0 \rightarrow 0.0$ \\
 \hline
 Weat-2 & $1.593  \rightarrow 1.615$ & $0.0 \rightarrow 0.0$ \\
 \hline
 Weat-3 & $-0.007  \rightarrow -0.006$ & $0.490 \rightarrow \textbf{0.492}$ \\
 \hline
 Weat-6 & $1.534  \rightarrow 1.605$ & $0.0 \rightarrow 0.0$ \\
 \hline
 Weat-7 & $1.484  \rightarrow \textbf{1.260}$ & $0.001 \rightarrow \textbf{0.006}$ \\
 \hline
 Weat-8 & $1.147  \rightarrow \textbf{0.672}$ & $0.008 \rightarrow \textbf{0.104}$ \\
 \hline
 Weat-9 & $0.507  \rightarrow \textbf{0.479}$ & $0.129 \rightarrow \textbf{0.137}$ \\
 \hline
 Weat-10 & $0.521  \rightarrow 0.552$ & $0.167 \rightarrow 0.149$ \\
 \hline
 Weat-11 & $0.830  \rightarrow 1.052$ & $0.022 \rightarrow 0.004$ \\
 \hline
 Weat-12 & $0.890  \rightarrow 1.085$ & $0.042 \rightarrow 0.017$ \\
 \hline
 Weat-13 & $0.271  \rightarrow \textbf{-0.029}$ & $0.249 \rightarrow \textbf{0.452}$ \\
 \hline
 Weat-14 & $0.683  \rightarrow \textbf{-0.040}$ & $0.098 \rightarrow \textbf{0.499}$ \\
 \hline
 Weat-15 & $0.919  \rightarrow 0.958$ & $0.036 \rightarrow 0.029$ \\
 \hline
 Weat-16 & $1.123  \rightarrow \textbf{0.984}$ & $0.010 \rightarrow \textbf{0.026}$ \\
 \hline
  \end{tabular}
  \caption{Fasttext WEAT results, arrow indicates  before to after mitigating bias}
\label{FasttextWEAT}

\end{center}
\end{minipage}\hfill
\begin{minipage}[t]{0.48\linewidth}
\centering
\medskip

    \begin{center}
\small
    \begin{tabular}{|c|c|c|}
    \hline
        WEAT list &  Effect size d &  Significance p \\
       \hline
    Weat-1 & $1.580  \rightarrow \textbf{1.556}$ & $0.0 \rightarrow 0.0$ \\
 \hline
 Weat-2 & $1.613  \rightarrow \textbf{1.598}$ & $0.0 \rightarrow 0.0$ \\
 \hline
 Weat-3 & $0.722  \rightarrow \textbf{0.708}$ & $0.001 \rightarrow 0.001$ \\
 \hline
 Weat-6 & $1.866  \rightarrow \textbf{1.840}$ & $0.0 \rightarrow 0.0$ \\
 \hline
 Weat-7 & $1.759  \rightarrow \textbf{0.946}$ & $0.0 \rightarrow \textbf{0.032}$ \\
 \hline
 Weat-8 & $1.339  \rightarrow \textbf{0.392}$ & $0.0025 \rightarrow \textbf{0.231}$ \\
 \hline
 Weat-9 & $1.552  \rightarrow \textbf{1.550}$ & $0.007 \rightarrow 0.007$ \\
 \hline
 Weat-10 & $0.239  \rightarrow \textbf{0.154}$ & $0.3329 \rightarrow \textbf{0.385}$ \\
 \hline
 Weat-11 & $1.433  \rightarrow \textbf{1.310}$ & $0.0 \rightarrow 0.0$ \\
 \hline
 Weat-12 & $1.560  \rightarrow 1.611$ & $0.0 \rightarrow 0.0$ \\
 \hline
 Weat-13 & $0.405  \rightarrow \textbf{0.113}$ & $0.185 \rightarrow \textbf{0.377}$ \\
 \hline
 Weat-14 & $1.417  \rightarrow \textbf{0.746}$ & $0.002 \rightarrow \textbf{0.077}$ \\
 \hline
 Weat-15 & $1.420  \rightarrow \textbf{0.763}$ & $0.001 \rightarrow \textbf{0.072}$ \\
 \hline
 Weat-16 & $1.338  \rightarrow \textbf{0.875}$ & $0.003 \rightarrow \textbf{0.046}$ \\
 \hline
  \end{tabular}
  \caption{COW-320 WEAT results, arrow indicates before to after mitigating bias}

\label{COW-320-WEAT}
\end{center}
\end{minipage}\hfill
\end{table}

\begin{table}[htb!]
\begin{minipage}[t]{0.48\linewidth}
\centering
\medskip
\small
   
\begin{center}

    \begin{tabular}{|c|c|c|}
    \hline
        WEAT list &  Effect size d &  Significance p \\
       \hline
    Weat-1 & $1.546  \rightarrow \textbf{1.530}$ & $0.0 \rightarrow 0.0$ \\
 \hline
 Weat-2 & $1.549  \rightarrow \textbf{1.538}$ & $0.0 \rightarrow 0.0$ \\
 \hline
 Weat-3 & $0.657  \rightarrow 0.657$ & $0.004 \rightarrow 0.004$ \\
 \hline
 Weat-6 & $1.771  \rightarrow \textbf{1.738}$ & $0.0 \rightarrow 0.0$ \\
 \hline
 Weat-7 & $1.713  \rightarrow \textbf{1.099}$ & $0.0 \rightarrow \textbf{0.014}$ \\
 \hline
 Weat-8 & $1.425  \rightarrow \textbf{0.506}$ & $0.001 \rightarrow \textbf{0.171}$ \\
 \hline
 Weat-9 & $1.388  \rightarrow \textbf{1.379}$ & $0.014 \rightarrow \textbf{0.015}$ \\
 \hline
 Weat-10 & $0.691  \rightarrow \textbf{0.659}$ & $0.097 \rightarrow \textbf{0.11}$ \\
 \hline
 Weat-11 & $1.399  \rightarrow \textbf{1.320}$ & $0.0 \rightarrow 0.0$ \\
 \hline
 Weat-12 & $1.526  \rightarrow 1.716$ & $0.0 \rightarrow 0.0$ \\
 \hline
 Weat-13 & $0.378  \rightarrow \textbf{0.081}$ & $0.202 \rightarrow \textbf{0.398}$ \\
 \hline
 Weat-14 & $1.334  \rightarrow \textbf{0.520}$ & $0.003 \rightarrow \textbf{0.169}$ \\
 \hline
 Weat-15 & $1.668  \rightarrow \textbf{1.115}$ & $0.001 \rightarrow \textbf{0.014}$ \\
 \hline
 Weat-16 & $1.46  \rightarrow \textbf{0.909}$ & $0.001 \rightarrow \textbf{0.038}$ \\
 \hline
  \end{tabular}
      \caption{COW-big WEAT results, arrow indicates  before to after mitigating bias}
\label{COW-big-weat}
\end{center}

\end{minipage}\hfill
\begin{minipage}[t]{0.48\linewidth}
\centering
\medskip
    \begin{center}
\small
   
\begin{tabular}{|c|c|c|}
    \hline
        WEAT list &  Effect size d &  Significance p \\
       \hline
    Weat-1 & $1.449  \rightarrow \textbf{1.427}$ & $0.0 \rightarrow 0.0$ \\
 \hline
 Weat-2 & $1.580  \rightarrow \textbf{1.564}$ & $0.0 \rightarrow 0.0$ \\
 \hline
 Weat-3 & $0.001  \rightarrow \textbf{-0.109}$ & $0.298 \rightarrow \textbf{0.456}$ \\
 \hline
 Weat-6 & $0.726  \rightarrow \textbf{0.578}$ & $0.084 \rightarrow \textbf{0.139}$ \\
 \hline
 Weat-7 & $1.451  \rightarrow \textbf{0.410}$ & $0.001 \rightarrow \textbf{0.215}$ \\
 \hline
 Weat-8 & $1.181  \rightarrow \textbf{0.271}$ & $0.007 \rightarrow \textbf{0.321}$ \\
 \hline
 Weat-9 & $1.287  \rightarrow 1.288$ & $0.013 \rightarrow \textbf{0.014}$ \\
 \hline
 Weat-10 & $0.070  \rightarrow \textbf{-0.110}$ & $0.449 \rightarrow 0.420$ \\
 \hline
 Weat-11 & $0.952  \rightarrow \textbf{0.893}$ & $0.009 \rightarrow \textbf{0.014}$ \\
 \hline
 Weat-12 & $1.436  \rightarrow \textbf{1.415}$ & $0.001 \rightarrow \textbf{0.002}$ \\
 \hline
 Weat-13 & $0.528  \rightarrow \textbf{0.401}$ & $0.106 \rightarrow \textbf{0.159}$ \\
 \hline
 Weat-14 & $1.315  \rightarrow 1.397$ & $0.002 \rightarrow 0.002$ \\
 \hline
 Weat-15 & $1.243  \rightarrow \textbf{1.107}$ & $0.004 \rightarrow \textbf{0.015}$ \\
 \hline
 Weat-16 & $1.199  \rightarrow 1.302$ & $0.005 \rightarrow 0.004$ \\
 \hline
  \end{tabular}
      \caption{Sonar 160 WEAT results, arrow indicates  before to after mitigating bias}
\label{sonar160weat}
\end{center}
\end{minipage}
\end{table}

\begin{table}[htb!]
   \begin{minipage}[t]{0.48\linewidth}
\centering
\medskip
\begin{center}
\small
    \begin{tabular}{|c|c|c|}
    \hline
        WEAT list &  Effect size d &  Significance p \\
       \hline
    Weat-1 & $1.413  \rightarrow \textbf{1.396}$ & $0.0 \rightarrow 0.0$ \\
 \hline
 Weat-2 & $1.572  \rightarrow \textbf{1.560}$ & $0.0 \rightarrow 0.0$ \\
 \hline
 Weat-3 & $-0.490  \rightarrow \textbf{-0.559}$ & $0.120 \rightarrow 0.070$ \\
 \hline
 Weat-6 & $0.528  \rightarrow \textbf{0.526}$ & $0.161 \rightarrow \textbf{0.163}$ \\
 \hline
 Weat-7 & $1.172  \rightarrow \textbf{0.643}$ & $0.011 \rightarrow \textbf{0.109}$ \\
 \hline
 Weat-8 & $0.995  \rightarrow \textbf{0.615}$ & $0.022 \rightarrow \textbf{0.128}$ \\
 \hline
 Weat-9 & $1.134  \rightarrow \textbf{1.133}$ & $0.021 \rightarrow \textbf{0.022}$ \\
 \hline
 Weat-10 & $-0.021  \rightarrow \textbf{-0.158}$ & $0.483 \rightarrow 0.384$ \\
 \hline
 Weat-11 & $0.727  \rightarrow \textbf{0.724}$ & $0.039 \rightarrow \textbf{0.041}$ \\
 \hline
 Weat-12 & $1.276  \rightarrow 1.343$ & $0.005 \rightarrow 0.003$ \\
 \hline
 Weat-13 & $0.445  \rightarrow 0.461$ & $0.140 \rightarrow 0.128$ \\
 \hline
 Weat-14 & $1.255  \rightarrow 1.523$ & $0.005 \rightarrow 0.001$ \\
 \hline
 Weat-15 & $1.092  \rightarrow \textbf{1.047}$ & $0.0134 \rightarrow \textbf{0.019}$ \\
 \hline
 Weat-16 & $1.150  \rightarrow 1.189$ & $0.009 \rightarrow 0.008$ \\
 \hline
  \end{tabular}
     \caption{Sonar 320 WEAT results, arrow indicates  before to after mitigating bias} 
\label{sonar320weat}
\end{center}
\end{minipage}\hfill
\begin{minipage}[t]{0.48\linewidth}
\centering
\medskip

    
\begin{center}
\small
    \begin{tabular}{|c|c|c|}
    \hline
        WEAT list &  Effect size d &  Significance p \\
       \hline
    Weat-1 & $1.624  \rightarrow \textbf{1.561}$ & $0.0 \rightarrow 0.0$ \\
 \hline
 Weat-2 & $1.538  \rightarrow \textbf{1.472}$ & $0.0 \rightarrow 0.0$ \\
 \hline
 Weat-3 & $0.493  \rightarrow \textbf{0.340}$ & $0.023 \rightarrow \textbf{0.094}$ \\
 \hline
 Weat-6 & $1.748  \rightarrow \textbf{1.721}$ & $0.0 \rightarrow 0.0$ \\
 \hline
 Weat-7 & $1.443  \rightarrow \textbf{1.161}$ & $0.001 \rightarrow \textbf{0.010}$ \\
 \hline
 Weat-8 & $0.766  \rightarrow 0.885$ & $0.071 \rightarrow 0.044$ \\
 \hline
 Weat-9 & $1.368  \rightarrow \textbf{1.366}$ & $0.009 \rightarrow 0.009$ \\
 \hline
 Weat-10 & $0.333  \rightarrow \textbf{0.084}$ & $0.274 \rightarrow \textbf{0.438}$ \\
 \hline
 Weat-11 & $1.482  \rightarrow \textbf{1.455}$ & $0.0 \rightarrow \textbf{0.001}$ \\
 \hline
 Weat-12 & $1.742  \rightarrow 1.821$ & $0.001 \rightarrow 0.0$ \\
 \hline
 Weat-13 & $0.537  \rightarrow \textbf{0.204}$ & $0.122 \rightarrow \textbf{0.306}$ \\
 \hline
 Weat-14 & $1.138  \rightarrow \textbf{0.790}$ & $0.011 \rightarrow \textbf{0.065}$ \\
 \hline
 Weat-15 & $1.416  \rightarrow \textbf{1.162}$ & $0.001 \rightarrow \textbf{0.008}$ \\
 \hline
 Weat-16 & $1.317  \rightarrow \textbf{1.241}$ & $0.002 \rightarrow \textbf{0.005}$ \\
 \hline
  \end{tabular}
     \caption{NLPL WEAT results, arrow indicates  before to after mitigating bias}
\label{nlplappendix}
\end{center}
\end{minipage}
\end{table}

\begin{table}[htb!]
    \begin{minipage}[t]{0.48\linewidth}
\centering
\medskip

\begin{center}
\small
    \begin{tabular}{|c|c|c|}
    \hline
        SEAT list &  Effect size d &  Significance p \\
       \hline
    SEAT-01 & $0.033  \rightarrow 0.0378$ & $0.32 \rightarrow \textbf{0.330} $ \\
 \hline
 SEAT-02 & $0.320  \rightarrow \textbf{0.315}$ & $0.01 \rightarrow 0.010 $ \\
 \hline
 SEAT-03 & $0.064  \rightarrow \textbf{0.055}$ & $0.25 \rightarrow \textbf{0.330} $ \\
 \hline
 SEAT-03b & $0.225  \rightarrow 0.228$ & $0.03 \rightarrow 0.010 $ \\
 \hline
 SEAT-04 & $0.075  \rightarrow 0.082$ & $0.25 \rightarrow 0.210 $ \\
 \hline
 SEAT-05 & $0.146  \rightarrow 0.157$ & $0.08 \rightarrow \textbf{0.100} $ \\
 \hline
 SEAT-05b & $0.345  \rightarrow 0.346$ & $0.0 \rightarrow 0.000 $ \\
 \hline
 SEAT-06 & $0.666  \rightarrow \textbf{0.614}$ & $0.01 \rightarrow 0.010 $ \\
 \hline
 SEAT-06b & $0.349  \rightarrow \textbf{0.294}$ & $0.02 \rightarrow \textbf{0.030} $ \\
 \hline
 SEAT-07 & $0.630  \rightarrow \textbf{0.588}$ & $0.01 \rightarrow 0.010 $ \\
 \hline
 SEAT-07b & $0.613  \rightarrow \textbf{0.605}$ & $0.01 \rightarrow 0.010 $ \\
 \hline
 SEAT-08 & $0.089  \rightarrow 0.114$ & $0.31 \rightarrow 0.290 $ \\
 \hline
 SEAT-08b & $0.286  \rightarrow 0.294$ & $0.06 \rightarrow \textbf{0.090} $ \\
 \hline
 SEAT-09 & $0.505  \rightarrow 0.507$ & $0.09 \rightarrow \textbf{0.110} $ \\
 \hline
 SEAT-10 & $0.606  \rightarrow \textbf{0.596}$ & $0.0 \rightarrow 0.000 $ \\
 \hline
  \end{tabular}
     \caption{BERTJe SEAT results, arrow indicates  before to after mitigating bias}
\label{bertjeseatappendix}
\end{center}
\end{minipage}\hfill
\begin{minipage}[t]{0.48\linewidth}
\centering
\medskip

   
\begin{center}
\small
    \begin{tabular}{|c|c|c|}
    \hline
        SEAT list &  BERTJe &  RobBERT \\ \hline
 SEAT-06 & $0.666^{**}  \rightarrow \textbf{0.614}^{**}$ & $0.191  \rightarrow \textbf{0.190}$ \\
 \hline
 SEAT-06b & $0.349^{*}  \rightarrow \textbf{0.294}^{*}$ & $0.552^{**}   \rightarrow 0.585^{**} $ \\
 \hline
 SEAT-07 & $0.630^{**}  \rightarrow \textbf{0.588}^{**}$ & $0.331^{*}  \rightarrow \textbf{0.252}$  \\
 \hline
 SEAT-07b & $0.613^{**}  \rightarrow \textbf{0.605}^{**}$ &$0.683^{**}  \rightarrow \textbf{0.669}^{**}$  \\
 \hline
 SEAT-08 & $0.089  \rightarrow 0.117$ &$0.080  \rightarrow 0.105$ \\
 \hline
 SEAT-08b & $0.286  \rightarrow 0.294$ & $0.157  \rightarrow \textbf{0.149}$ \\
 \hline

\multicolumn{3}{l}{Arrow indicates before to after mitigating bias; * indicates}\\
\multicolumn{3}{l}{ significant at 0.05, ** significant at 0.01.}
 \end{tabular}
      \caption{SEAT results on gender related test.}
\label{SEAT_GENDER}
\end{center}
\end{minipage}
\end{table}

\begin{table}[htb!]
   
 \begin{center}
\small
    \begin{tabular}{|c|c|c|}
    \hline
        SEAT list &  Effect size d &  Significance p \\
       \hline
    SEAT-01 & $0.205  \rightarrow 0.211$ & $0.000 \rightarrow 0.000 $ \\
 \hline
 SEAT-02 & $0.131  \rightarrow 0.134$ & $0.040 \rightarrow 0.040 $ \\
 \hline
 SEAT-03 & $0.119  \rightarrow \textbf{0.093}$ & $0.090 \rightarrow \textbf{0.200} $ \\
 \hline
 SEAT-03b & $0.025  \rightarrow 0.032$ & $0.410 \rightarrow 0.260 $ \\
 \hline
 SEAT-04 & $0.071  \rightarrow \textbf{0.067}$ & $0.320 \rightarrow 0.220 $ \\
 \hline
 SEAT-05 & $0.679  \rightarrow \textbf{0.676}$ & $0.010 \rightarrow 0.010 $ \\
 \hline
 SEAT-05b & $0.098  \rightarrow 0.101$ & $0.140 \rightarrow 0.10 $ \\
 \hline
 SEAT-06 & $0.191  \rightarrow \textbf{0.190}$ & $0.120 \rightarrow \textbf{0.180} $ \\
 \hline
 SEAT-06b & $0.552  \rightarrow 0.585$ & $0.000 \rightarrow 0.000 $ \\
 \hline
 SEAT-07 & $0.331  \rightarrow \textbf{0.252}$ & $0.030 \rightarrow \textbf{0.080} $ \\
 \hline
 SEAT-07b & $0.683  \rightarrow \textbf{0.669}$ & $0.000 \rightarrow 0.000 $ \\
 \hline
 SEAT-08 & $0.080  \rightarrow 0.105$ & $0.33 \rightarrow 0.310 $ \\
 \hline
 SEAT-08b & $0.157  \rightarrow \textbf{0.149}$ & $0.260 \rightarrow 0.260 $ \\
 \hline
 SEAT-09 & $0.634  \rightarrow 0.638$ & $0.040 \rightarrow 0.030 $ \\
 \hline
 SEAT-10 & $0.642  \rightarrow \textbf{0.637}$ & $0.010 \rightarrow 0.010 $ \\
 \hline
  \end{tabular}
     \caption{RobBERT SEAT results, arrow indicates  before to after mitigating bias}
\label{robbertseatappendix}
\end{center}
\end{table}

\begin{table*}[htb!]
    \begin{center}
    \small
\begin{tabular}{|l|l|l|l|l|}
\hline
Test   & M                         & F                        & A                  & B                    \\ \hline
 Weat-1   & flowers                   & insects                  & pleasant           & unpleasant           \\ \hline
 Weat-2   & instruments               & weapons                  & pleasant           & unpleasant           \\ \hline
 Weat-3   & european\_american\_names & african\_american\_names & pleasant           & unpleasant           \\ \hline
 Weat-6   & male\_names               & female\_names            & career             & family               \\ \hline
 Weat-7   & math                      & arts                     & male\_terms        & female\_terms        \\ \hline
 Weat-8   & science                   & arts                     & male\_terms        & female\_terms        \\ \hline
 Weat-9   & mental\_disease           & physical\_disease        & temporary          & permanent            \\ \hline
 Weat-10  & young\_peoples\_names     & old\_peoples\_names      & pleasant           & unpleasant           \\ \hline
 Weat-11  & male\_terms               & female\_terms            & career             & family               \\ \hline
 Weat-12  & career                    & family                   & male\_terms        & female\_terms        \\ \hline
 Weat-13  & math                      & arts                     & male\_names\_dutch & female\_names\_dutch \\ \hline
 Weat-14  & science                   & arts                     & male\_names\_dutch & female\_names\_dutch \\ \hline
 Weat-15 & male\_names\_dutch        & female\_names\_dutch     & career             & family               \\ \hline
 Weat-16 & career                    & family                   & male\_names\_dutch & female\_names\_dutch \\ \hline
\end{tabular}

\caption{Weat tests lists categories}
\label{weat_categories}
\end{center}
\end{table*}

\end{document}